\documentclass[conference]{IEEEtran}
\IEEEoverridecommandlockouts
\usepackage{cite}
\usepackage{amsmath,amssymb,amsfonts}
\usepackage[linesnumbered,ruled,boxed,norelsize,vlined,commentsnumbered]{algorithm2e}

\usepackage{multirow} 
\usepackage{graphicx}
\usepackage{textcomp}
\usepackage{adjustbox}
\usepackage{xcolor}
\usepackage{booktabs}
\usepackage[inkscapelatex=false]{svg}

\def\BibTeX{{\rm B\kern-.05em{\sc i\kern-.025em b}\kern-.08em
    T\kern-.1667em\lower.7ex\hbox{E}\kern-.125emX}}
\begin{document}

\title{Cognitive-Inspired Hierarchical Attention Fusion With Visual and Textual for Cross-Domain Sequential Recommendation\\

\thanks{$^{\dag}$Corresponding author}\thanks{This work was supported by the National Natural Science Foundation of China (No. 62471405, 62331003, 62301451), Suzhou Basic Research Program (SYG202316) and XJTLU REF-22-01-010, XJTLU AI University Research Centre, Jiangsu Province Engineering Research Centre of Data Science and Cognitive Computation at XJTLU and SIP AI innovation platform (YZCXPT2022103).}
}

% \author{

% \IEEEauthorblockN{1\textsuperscript{st} Wangyu Wu}
% \IEEEauthorblockA{\textit{Xi’an Jiaotong-Liverpool University} \\
% \textit{The University of Liverpool}\\
% Suzhou, China \\
% v11dryad@foxmail.com}\\
% % \and
% \IEEEauthorblockN{3\textsuperscript{nd} Given Name Surname}
% email address or ORCID}\\
% \and
% \IEEEauthorblockN{2\textsuperscript{rd} Xiaowei Huang}
% \IEEEauthorblockA{\textit{The University of Liverpool} \\
% Liverpool, UK \\
% xiaowei.huang@liverpool.ac.uk}\\\\
% % \and
% \IEEEauthorblockN{4\textsuperscript{th} Fei Ma$^{{\dag}}$}
% \IEEEauthorblockA{\textit{Xi’an Jiaotong-Liverpool University} \\
% Suzhou, China \\
% fei.ma@xjtlu.edu.cn}\\
%  \and
% \IEEEauthorblockN{5\textsuperscript{th} Jimin Xiao$^{{\dag}}$}
% \IEEEauthorblockA{\textit{Xi’an Jiaotong-Liverpool University} \\
%  Suzhou, China \\
%  jimin.xiao@xjtlu.edu.cn}
% }

\author{Wangyu Wu\textsuperscript{1,2} \quad
Zhenhong Chen\textsuperscript{3} \quad Xianglin Qiu\textsuperscript{1,2} \quad
Siqi Song\textsuperscript{1,2}  \\
Xiaowei Huang\textsuperscript{2} \quad
Fei Ma\textsuperscript{1*} \quad
Jimin Xiao\textsuperscript{1}\thanks{Corresponding author} \\
\textsuperscript{1} Xi’an Jiaotong-Liverpool University \quad  \textsuperscript{2} The University of Liverpool \\
\textsuperscript{3} Microsoft \\
\tt\small \{Wangyu.wu22, Xianglin.Qiu20, Siqi.Song22\}@student.xjtlu.edu.cn \\  \tt\small \{fei.ma, jimin.xiao\}@xjtlu.edu.cn
}

\maketitle

\begin{abstract}
Cross-Domain Sequential Recommendation (CDSR) predicts user behavior by leveraging historical interactions across multiple domains, focusing on modeling cross-domain preferences through intra- and inter-sequence item relationships. Inspired by human cognitive processes, we propose Hierarchical Attention Fusion of Visual and Textual Representations (HAF-VT), a novel approach integrating visual and textual data to enhance cognitive modeling. Using the frozen CLIP model, we generate image and text embeddings, enriching item representations with multimodal data. A hierarchical attention mechanism jointly learns single-domain and cross-domain preferences, mimicking human information integration. Evaluated on four e-commerce datasets, HAF-VT outperforms existing methods in capturing cross-domain user interests, bridging cognitive principles with computational models and highlighting the role of multimodal data in sequential decision-making.

\end{abstract}

\section{Introduction}\label{sec:intro}
Sequential Recommendation (SR) has emerged as a prominent approach for modeling dynamic user preferences by analyzing historical interaction sequences~\cite{markovsr,transmarkov,markov,narm}. While SR focuses on predicting the next item of interest within a single domain, it often suffers from data sparsity and domain bias, limiting its ability to capture the full spectrum of user behavior. To address these limitations, Cross-Domain Sequential Recommendation (CDSR)~\cite{pinet,Zhuangict,dagcn,kddsemi,cao2022contrastive,recguru} has been proposed, leveraging information from multiple domains to enhance recommendation accuracy and provide a more comprehensive understanding of user preferences.

\begin{figure}[t]
\centering
\includegraphics[width=0.9\linewidth]{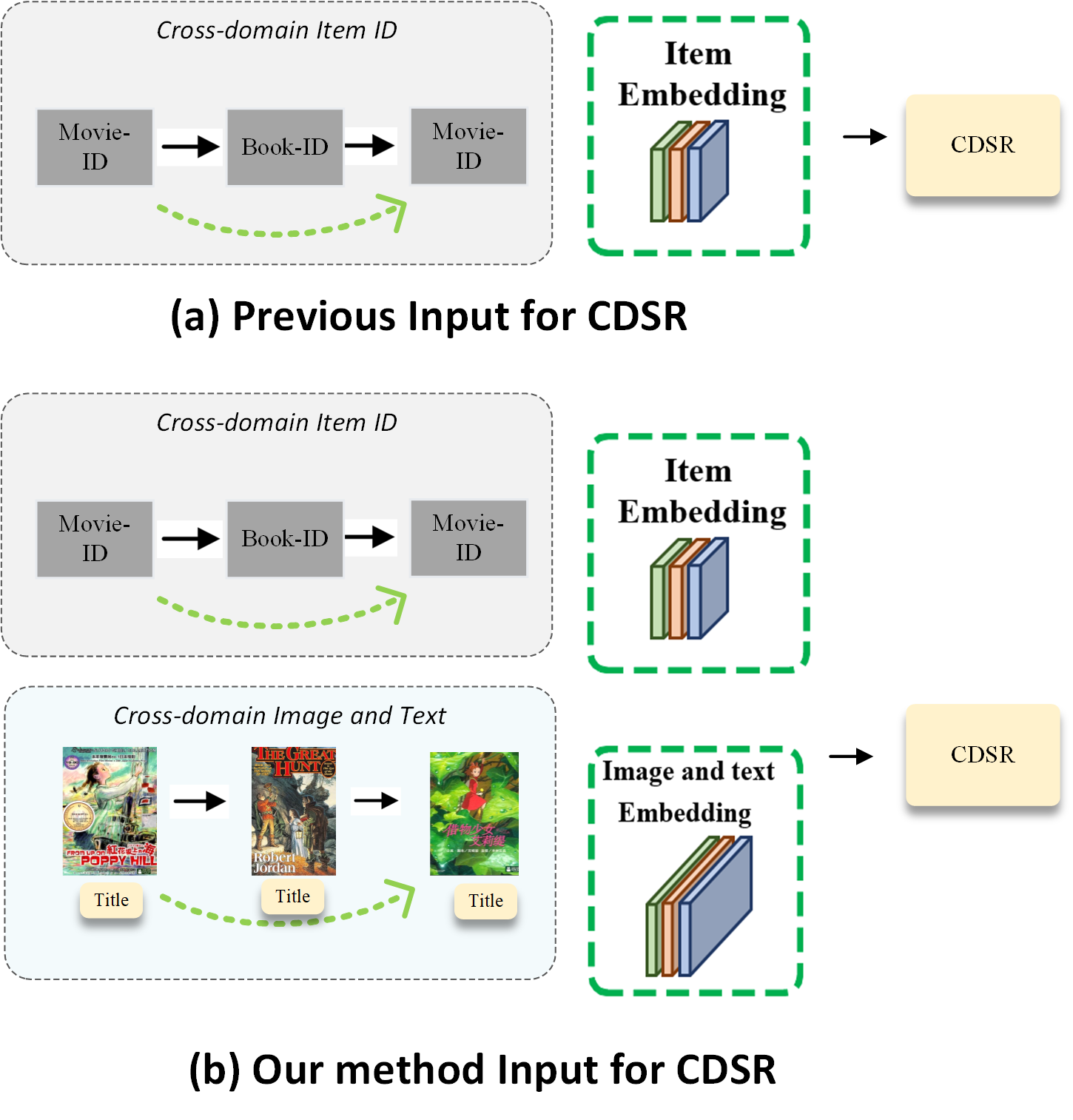}
% \vspace{-0.3cm} 
\caption{(a) In the traditional CDSR framework, the input consists solely of item ID features. (b) In our HAF-VT framework, we incorporate additional image and title information to complement the existing item features. Therefore, our method enriches item representations.}
\label{fig:idea}
% \vspace{-0.3cm} 
\end{figure}

Early efforts in CDSR, such as $\pi$-Net~\cite{pinet}, model item interaction sequences within a single domain and transfer learned representations to other domains through gated transfer modules. Subsequent work, like MIFN~\cite{mifn}, incorporates external knowledge graphs to strengthen cross-domain connections. However, these approaches primarily focus on intra-sequence relationships within individual domains, often neglecting inter-sequence relationships across domains. This oversight leads to persistent domain bias and suboptimal performance. 

Moreover, existing CDSR methods fail to fully exploit the rich visual information associated with items. Human decision-making is inherently multimodal, with visual cues playing a critical role in shaping user preferences~\cite{hegarty2011cognitive}. For instance, users often form initial impressions of items based on their visual appearance, even when textual descriptions are similar. This observation aligns with cognitive theories emphasizing the integration of multimodal stimuli in preference formation~\cite{shams2008benefits}. To bridge this gap, we propose \textbf{Hierarchical Attention Fusion of Visual and Textual Representations (HAF-VT)}, a novel framework that integrates visual and textual embeddings to enrich item representations and model cross-domain user preferences. By leveraging the frozen CLIP model~\cite{radford2021learning}, we generate aligned image and text embeddings, capturing both intra- and inter-sequence relationships. Our hierarchical attention mechanism jointly learns single-domain and cross-domain preferences, mimicking the human ability to integrate information across contexts.

\textbf{Our main contributions are summarized as follows:}
\begin{itemize}[leftmargin=1em]
    \item We introduce the CDSR framework that integrates visual, textual, and item embeddings, enriching item representations and enhancing recommendation performance through multimodal cognitive modeling.
    \item We propose a hierarchical attention mechanism that captures intra-sequence and inter-sequence relationships, enabling more accurate next-item prediction by aligning with human cognitive processes.
    \item We conduct extensive experiments on four reorganized e-commerce datasets, demonstrating that HAF-VT outperforms state-of-the-art baselines across all evaluation metrics.
\end{itemize}

\begin{figure*}[t] 
\begin{center}
   \includegraphics[width=1\linewidth]{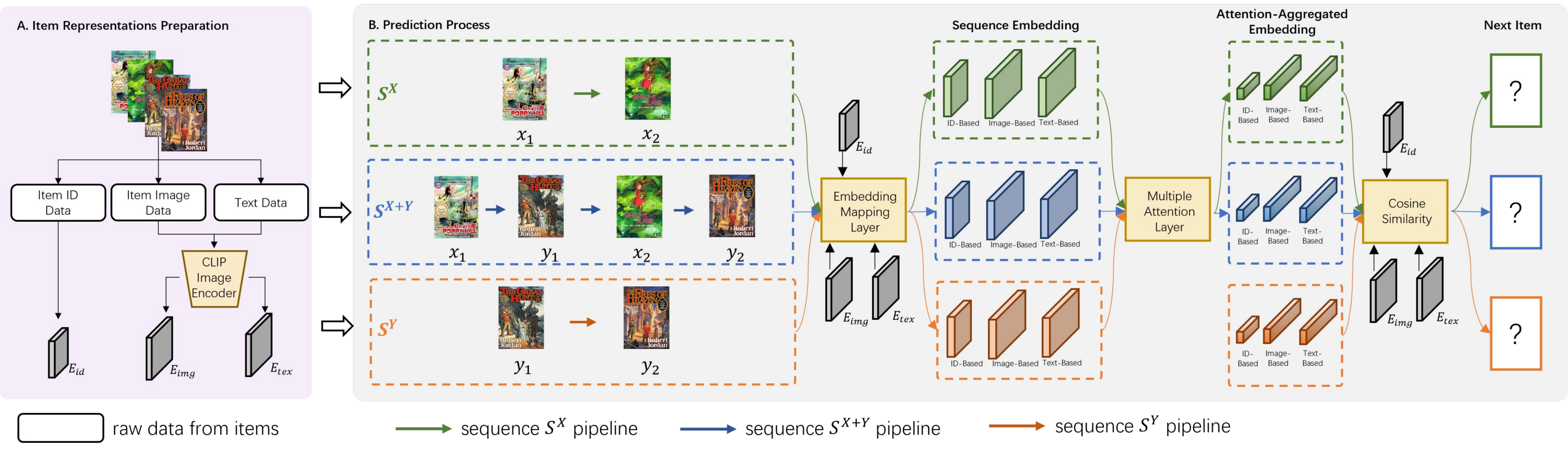}

   \caption{The overview of our proposed HAF-VT is as follows. Firstly, the Feature Preparation Process embeds all items from domains $X$ and $Y$ using a learnable item matrix for ID-based embeddings $E_{id}$, a frozen CLIP image encoder for image-based embeddings $E_{img}$, and a text encoder for text-based embeddings $E_{tex}$. Then, the input sequence $\mathcal{S}$ is forwarded to the Image Fusion Prediction Process. $\mathcal{S}$ consists of sub-sequences $S^X$, $S^Y$, and $S^{X+Y}$, which are processed through an embedding mapping layer to generate item ID-based, image-based, and text-based sequence embeddings. These embeddings are passed through multiple attention layers to capture both intra- and inter-sequence relationships, producing attention-aggregated embeddings. Finally, cosine similarity is applied between these aggregated embeddings and $E_{id}$, $E_{img}$, and $E_{tex}$ to predict the next item.}
    \label{fig:framework}
\end{center}
% \vspace{-0.3cm}
\end{figure*}

\section{Related Work}

\subsection{Models for Cognitive Science}
Recent advancements in computer vision and machine learning~\cite{cao2024slcf,cao2025ocsop,cao2025swasop,cao2024diffssc,yu2025crisp,yu2025prnet,yu2024scnet,li2024towards,li2024distinct,li2024towardsv2} have demonstrated significant potential for addressing key research questions in cognitive science~\cite{zhou2025image,zhou2024smart,yang2025magic,yang2025multimodal,li2023diverse,licore,liegoprivacy}. A growing body of work has focused on image similarity and categorization~\cite{yao2024cracknex,yao2024event,zhou2024two, chen2024adaptive, qiu2024tfb,wu2025spiking}, which, while related to the phenomenon of naming, does not directly address it. These studies collectively reveal that deep visual representations, particularly those derived from the final layers of neural networks, align closely with human perceptual judgments. This alignment is often measured by comparing human similarity ratings with model-generated similarity scores, highlighting the ability of deep learning models to capture human-like visual representations. More directly relevant to the present work are studies by~\cite{wu2025generative,gualdoni2023s, gunther2023vispa,wu2024prompt}, which leverage deep learning-based computer vision models to construct visual semantic representations for extensive vocabularies. These approaches bridge the gap between visual and linguistic domains, offering insights into how visual information can be mapped to semantic structures. Recent research has also begun to explore cognitive modeling using the CLIP (Contrastive Language–Image Pretraining) framework. The integration of self-supervised learning with vision-language pretraining has spurred the development of numerous CLIP-based models, each aiming to enhance the alignment between visual and textual data~\cite{cha2023learning}. These models extend the original CLIP architecture by introducing novel techniques to improve representation quality and strengthen the connection between images and text.

Despite these advancements, existing methods often fail to fully integrate visual and textual information to enhance model perception. In contrast, our approach employs a Hierarchical Attention mechanism to effectively capture and combine information from multiple feature sources, thereby improving the overall performance and alignment with human cognitive processes.

\subsection{Sequential Recommendation}
Sequential recommendation~\cite{SRs} serves as the core framework of our approach, where user behavior is modeled as a sequence of time-sensitive items with the goal of predicting the likelihood of the next item in the sequence. Initially, Markov chains were utilized to capture the sequential dependencies within the item sequences, as seen in FPMC~\cite{rendle2010factorizing}.  
To capture more complex interactions between historical items, deep learning techniques such as recurrent neural networks~\cite{GRU, LSTM}, convolutional neural networks~\cite{CNN}, and attention mechanisms~\cite{vaswani2017attention} have been integrated into recommender systems~\cite{gru4rec, DIEN, Caser, kang2018self, DIN}. In addition, SURGE~\cite{SURGE} uses metric learning to reduce the complexity of the item sequences. More recently, works like DFAR~\cite{lin2023dual} and DCN~\cite{lin2022dual} have explored capturing deeper, more intricate relationships within sequential recommendation data. In this work, we extend these ideas by applying cross-domain learning to sequential recommendation models, facilitating knowledge transfer between multiple domains.

\subsection{Cross-Domain Recommendation} Cross-domain recommender systems~\cite{CDR} have emerged as a promising approach to address the challenges of sparse data and cold-start issues typically faced in sequential recommendation tasks~\cite{wuimgfu,cai2024relation}. Early models in this area were grounded in single-domain recommendation frameworks, under the assumption that auxiliary user behaviors from different domains could contribute to improved user modeling in the target domain~\cite{singh_relational_2008}. Many popular methods rely on transfer learning~\cite{Transfer}, which involves transferring user or item embeddings from the source domain to the target domain to enhance the modeling process. Notable examples of such approaches include MiNet~\cite{MiNet}, CoNet~\cite{CoNet}, and itemCST~\cite{itemCST}, among others. However, in practice, industrial platforms typically aim to improve all product domains simultaneously, rather than exclusively enhancing the target domain without considering the source domain. This has led to the development of dual learning~\cite{qiu2025easytime, qiu2025duet}, which enables mutual improvements across both the source and target domains, and has already been successfully applied in cross-domain recommender systems. Despite these advancements, many existing methods fail to account for the role of visual perceptual features and do not integrate multimodal fusion with attention mechanisms, which could significantly enhance performance.

\section{Methodology}
\label{sec:method}

\subsection{Problem Formulation}
In the CDSR task, user interaction sequences occur in domain $X$ and domain $Y$, with their item sets denoted as $\mathcal{X}$ and $\mathcal{Y}$, respectively. Let $\mathcal{S}$ represent the overall interaction sequence of the user in chronological order, which consists of three sub-sequences ${(S^X, S^Y, S^{X+Y}) \in \mathcal{S}}$. Specifically, $S^X = {\left [ x_1, x_2,\dots , x_ {\left | S^X \right |}  \right ]}, {x\in \mathcal{X}}$ and $S^Y = {\left [ y_1, y_2,\dots , y_ {\left | S^Y \right |}  \right ]}, {y\in \mathcal{Y}}$ are the interaction sequences within each domain, where ${\left | \cdot  \right | }$ denotes the total item number. Additionally, ${S^{X+Y} = {\left [ x_1, y_1, x_2,\dots , x_ {\left | S^X \right |}, \dots ,y_ {\left | S^Y \right |}  \right ]}}$ represents the merged sequence by combining $S^X$ and $S^Y$. In general, the goal of the CDSR task is to predict the probabilities of  candidate items across both domains and select the item with the highest probability as the next  recommendation.

\subsection{Overall framework}
As illustrated in Fig.~\ref{fig:framework}, the proposed HAF-VT framework is designed to model cross-domain sequential preferences through a hierarchical cognitive architecture. The framework begins with Item Representation Encoding, where items from domains $X$ and $Y$ are processed through parallel pathways to construct distinct representations. A learnable semantic embedding matrix $E_{id}$ is initialized to capture identity-specific features, while a frozen CLIP encoder generates perceptual embeddings $E_{img}$ and conceptual embeddings $E_{tex}$ from item images and textual descriptions, respectively. This multimodal encoding stage simulates the human capacity for integrating visual, semantic, and linguistic information.

The framework then proceeds to the Preference Simulation stage, where behavioral sequences $\mathcal{S}$ are processed. Each sequence $\mathcal{S}$ consists of three sub-sequences: $S^X$, $S^Y$, and $S^{X+Y}$, representing within-domain and cross-domain interactions. The embedding layer retrieves and restructures the multimodal embeddings ($E_{id}$, $E_{img}$, and $E_{tex}$) according to the sequence order, generating dynamic cognitive states for each sub-sequence. These states are further processed through hierarchical attention layers to model both intra-sequence and inter-sequence relationships, simulating the interplay between local working memory and global contextual integration.

Finally, the framework performs Decision Generation by comparing the attention-aggregated sequence embeddings with the original multimodal embeddings ($E_{id}$, $E_{img}$, and $E_{tex}$) using cosine similarity. This process mimics human decision-making by combining evidence from multiple sensory modalities to predict the next item in the sequence.

\subsection{Visual and Textual Feature Integration}
\label{sec:Image}
We enhance user preferences by integrating multimodal features (image and textual) into the CDSR framework. This subsection provides a detailed explanation of feature preparation and sequence representation combined with image and textual features within a single domain, and applies this approach to domain $X$, domain $Y$, and domain $X+Y$.

\textit{1) Item Presentations Preparation with Image Feature:} 
First, we prepare the feature representations for all items. As shown on the left in Fig.~\ref{fig:framework}, we construct a learnable item matrix based on item IDs for all items in domain $X$ and $Y$, denoted as $E_{id} \in \mathbb{R}^{(|\mathcal{X}|+|\mathcal{Y}|) \times q}$. Here, $|\mathcal{X}|+|\mathcal{Y}|$ represents the total number of items from domains $X$ and $Y$, and $q$ is the learnable item embedding dimension. Simultaneously, we employ a pre-trained and frozen CLIP model to generate image embeddings for each item. These embeddings are used to form image matrix $E_{img} \in \mathbb{R}^{(|\mathcal{X}|+|\mathcal{Y}|) \times e}$, where $e$ is the image embedding dimension. These two matrices serve as the base for generating embeddings for sequences.

\textit{2) Item Presentations Preparation with Textual Feature:} 
To further enrich the item representations, we leverage textual information from item titles using the same frozen CLIP model. For each item, the CLIP text encoder generates textual embeddings based on the item's title, forming the textual matrix $E_{tex} \in \mathbb{R}^{(|\mathcal{X}|+|\mathcal{Y}|) \times e}$. This textual representation captures semantic information that complements the visual and ID-based features, enabling a more comprehensive understanding of user preferences.

\textit{3) Enhanced Similarity Score with Multimodal Features:}
We first generate a sequence representation to represent the user interaction sequence, which is then used to calculate the similarity with items for predicting the next item in the Prediction Process. As shown on the right in Fig.~\ref{fig:framework}, we propose an embedding layer to process the input sequences $\mathcal{S}$, which consists of three sub-sequences $S^X$, $S^Y$, and $S^{X+Y}$. These sequences include item IDs, images, and textual data. Accordingly, the embedding layer produces item ID-based, image-based, and textual-based embeddings for each input sequence. 

Once we obtain $E_{id}$, $E_{img}$, and $E_{tex}$, the item ID-based sequence embeddings $F_{id} \in \mathbb{R}^{|\mathcal{S}| \times q}$ for $\mathcal{S}$ can be generated by placing the appropriate embedding from $E_{id}$ in the order of the items in the sequence $\mathcal{S}$. Here, $|\mathcal{S}|$ denotes the total item number in $\mathcal{S}$. Similarly, the image-based sequence embeddings $F_{img} \in \mathbb{R}^{|\mathcal{S}| \times e}$ and textual-based sequence embeddings $F_{tex} \in \mathbb{R}^{|\mathcal{S}| \times e}$ are produced from $E_{img}$ and $E_{tex}$, respectively, based on the same sequence. 

For simplicity, we take the image-based sequence embeddings $F_{img}$ as an example to illustrate how HAF-VT obtains the next item prediction. We employ an Attention Layer to get the enhanced sequence image-based embeddings $H_{img} \in \mathbb{R}^{|\mathcal{S}| \times e}$ as follows:

\begin{equation}
    H_{img} = Attention(F_{img}),
\end{equation}

Afterward, we extract the last embedding vector $h_{img} \in \mathbb{R}^{1 \times e}$ from $H_{img}$ as the sequence representation, which serves as the attention-aggregated embedding of the sequence. This vector captures the user preferences, specifically focusing on the most recent interaction within the sequence. Then, $h_{img}$ is compared against $E_{img}$ using cosine similarity. In this way, the alignment between user preferences and the embedding of each item across all domains is assessed. The similarity score is computed as follows:

\begin{equation} 
\begin{aligned}\label{eq:cos}
Sim(h_{img}, E_{img}) = \frac{h_{img} \cdot E_{img}^{T}}{\|h_{img}\|\|E_{img}^T\|},
\end{aligned}
\end{equation}

where $T$ denotes the transpose function. Here, a higher similarity score indicates that the sequence preference is more aligned with this item. In the same manner, we can obtain the item ID-based sequence representative vector $h_{id} \in \mathbb{R}^{1 \times q}$ and textual-based sequence representative vector $h_{tex} \in \mathbb{R}^{1 \times e}$, along with their respective similarity scores. In the next section, we will introduce how the multiple attention layer is used to fuse ID-based, image-based, and textual-based predictions for CDSR.  
\begin{figure}[t]
\centering
\includegraphics[width=1.0\linewidth]{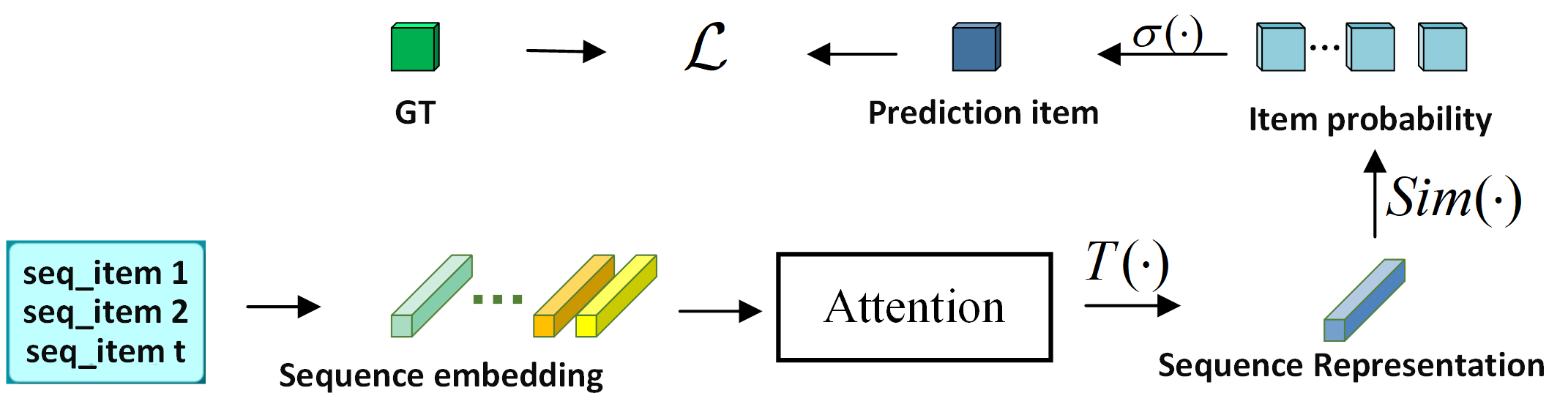}
% \vspace{-0.3cm} 
\caption{The process of transforming a user sequence into a representation sequence. Initially, the user’s sequence of interactions is converted into an embedded representation. This sequence is then passed through an attention layer to capture both intra- and inter-sequence relationships. Finally, the attention-aggregated sequence representations are compared with item embeddings, and the item with the highest similarity score is selected as the predicted next item.}
\label{fig:detail}
\end{figure}
\subsection{Hierarchical Attention in Preference Formation}
\label{sec:Attention}

In traditional sequential recommendation methods~\cite{gru4rec,sasrec}, behavioral sequences from domains $X$ and $Y$ are often processed indiscriminately, leading to potential dominance by the domain with higher behavioral frequency. This imbalance can skew the representation of user preferences. To address this issue, we propose a hierarchical attentional framework that separately processes sequences $S^X$, $S^Y$, and $S^{X+Y}$, ensuring balanced representation across domains. Our approach integrates item identity (ID), visual features, and textual embeddings to model user preferences more comprehensively. As illustrated in Fig.~\ref{fig:detail}, to model the interactions within each sequence more effectively, we incorporate a self-attention mechanism with query, key, and value (QKV) pairs. The attention mechanism allows the model to focus on relevant parts of the sequence when making predictions, capturing both local and global dependencies within the sequence.

The attention mechanism computes the output as a weighted sum of the values, where the weights are determined by the similarity between the query and the keys:

\begin{equation}
\text{Attention}(Q, K, V) = \text{softmax}\left(\frac{QK^T}{\sqrt{d_k}}\right) V,
\end{equation}
where $Q$, $K$, and $V$ are the query, key, and value matrices, respectively, and $d_k$ is the dimension of the key vectors. This mechanism is applied within each domain's sequence processing to capture both intra-domain and cross-domain relationships more effectively.

The framework generates nine sequence representations through multiple attention layers: $h^{X}_{id}$, $h^{X}_{img}$, $h^{X}_{tex}$, $h^{Y}_{id}$, $h^{Y}_{img}$, $h^{Y}_{tex}$, $h^{X+Y}_{id}$, $h^{X+Y}_{img}$, and $h^{X+Y}_{tex}$. These representations capture both intra-domain and cross-domain user preferences. For domain $X$, the prediction probability for the next item $x_{t+1}$ in sequence $S^X$ is computed as follows:

\begin{equation}
\small
\mathrm{P}^X_{id}(x_{t+1} \mid \mathcal{S}) = \text{softmax}\left(\text{Sim}(h_{id}^X, E_{id}^X)\right), \quad x_t \in S^X,
\end{equation}
where $x_{t+1}$ denotes the predicted next item, $h_{id}^X$ is the ID-based sequence representation, and $E_{id}^X$ is the ID embedding matrix for domain $X$. Similarly, the prediction probabilities based on visual and textual features are calculated as:

\begin{equation}
\small
\mathrm{P}^X_{img}(x_{t+1} \mid \mathcal{S}) = \text{softmax}\left(\text{Sim}(h_{img}^X, E_{img}^X)\right), \quad x_t \in S^X,
\end{equation}
\begin{equation}
\small
\mathrm{P}^X_{tex}(x_{t+1} \mid \mathcal{S}) = \text{softmax}\left(\text{Sim}(h_{tex}^X, E_{tex}^X)\right), \quad x_t \in S^X,
\end{equation}
where $h_{img}^X$, $E_{img}^X$, $h_{tex}^X$, and $E_{tex}^X$ represent the image-based sequence representation, image embedding matrix, text-based sequence representation, and text embedding matrix, respectively. The final prediction probability for the next item combines ID, image, and text-based predictions:

\begin{align}
\mathrm{P}^X(x_{t+1} \mid \mathcal{S}) &= \alpha \mathrm{P}^X_{id}(x_{t+1} \mid \mathcal{S}) \nonumber \\
&\quad + \beta \mathrm{P}^X_{img}(x_{t+1} \mid \mathcal{S}) \nonumber \\
&\quad + (1-\alpha-\beta) \mathrm{P}^X_{tex}(x_{t+1} \mid \mathcal{S}),
\end{align}

where $\alpha$ and $\beta$ are weighting factors that balance the contributions of ID, visual, and textual features. This formulation captures structural, perceptual, and semantic aspects of user preferences.

To optimize the framework, we minimize the following loss function for domain $X$:

\begin{equation}
\small
\mathcal{L}^X = \sum_{x_t \in S^X} -\log \mathrm{P}^X(x_{t+1} \mid \mathcal{S}).
\label{sec_math_eq:single}
\end{equation}

The same method is applied to domains $Y$ and $X+Y$, yielding loss terms $\mathcal{L}^Y$ and $\mathcal{L}^{X+Y}$. The final loss is a weighted combination of these terms:

\begin{equation}
\mathcal{L} = \mathcal{L}^{X} + \lambda_{1}\mathcal{L}^{Y} + \lambda_{2}\mathcal{L}^{X+Y},
\end{equation}
where $\lambda_1$ and $\lambda_2$ control the relative importance of each domain.

During evaluation, we aggregate predictions from all domains to recommend the next item. Let $X$ be the target domain and $Y$ the source domain. The combined prediction score for item $x_i$ is computed as:

\begin{equation}
\small
\mathrm{P}(x_i|S) = \mathcal{P}^{X}(x_i|S) + \lambda_{1} \mathcal{P}^{Y}(x_i|S) + \lambda_{2} \mathcal{P}^{X+Y}(x_i|S).
\end{equation}

The recommended item is selected by maximizing the prediction score within the target domain $X$:

\begin{equation}
\mathrm{argmax}_{x_i \in \mathcal{X}} \mathrm{P}(x_i|S).
\end{equation}

This process allows the model to incorporate cross-domain information effectively, leveraging the predictions from both the target domain and the source domain. By weighting the contributions from each domain, the model can refine its recommendation, taking into account the richer information that might come from the source domain while focusing on the target domain for the final prediction.

The weighting factors $\lambda_1$ and $\lambda_2$ allow the model to control how much influence the source domain ($Y$) and the combined domain ($X+Y$) have on the recommendation for the target domain. This flexibility in adjusting the weights is crucial for handling different types of datasets or domain-specific characteristics, ensuring that the model adapts to varying degrees of relevance between domains.

Finally, by selecting the item with the highest prediction score from the target domain, the model aims to provide the most contextually appropriate recommendation, which is critical for real-world recommendation systems that need to balance both accuracy and relevance across multiple domains.

\section{Experiments}
\label{sec:Experiments}

In this section, we introduce the experimental setup, providing detailed descriptions of the dataset, evaluation metrics, and implementation specifics. We then compare our method with state-of-the-art approaches on the Amazon dataset~\cite{wei2021contrastive}. Finally, we conduct ablation studies to validate the effectiveness of the proposed method.

\textbf{Dataset and Evaluated Metric.} Our experiments are conducted on the Amazon dataset~\cite{wei2021contrastive} to construct the CDSR scenarios. Following previous works~\cite{pinet, mifn}, we select four domains to generate two CDSR scenarios for our experiments: ``Food-Kitchen'' and ``Movie-Book''. We first extract users who have interactions in both domains. Then we filter out users and items with fewer than 10 interactions. Additionally, to meet the sequential constraints, we retain cross-domain interaction sequences that contain at least three items from each domain. In the training/validation/test split, the latest interaction sequences are equally divided into the validation and test sets, while the remaining interaction sequences are used for the training set. The statistics of our refined datasets for the CDSR scenarios are summarized in Table \ref{sec_exp_tab:dataset}. We adopt Mean Reciprocal Rank (MRR)~\cite{mrr} and Normalized Discounted Cumulative Gain (NDCG@)\{5, 10\}~\cite{ndcg} as evaluation metrics to assess the performance of our model. These metrics are widely used in recommendation systems.

\begin{table}[t]
\centering
\caption{Statistics of two CDSR scenarios.}
\vspace{0.3cm}
\setlength{\tabcolsep}{5pt}
\resizebox{8.5cm}{!}{
\begin{tabular}{cccccc}
\toprule
\textbf{Scenarios}  &\textbf{\#Items} &\textbf{\#Train} &\textbf{\#Valid}  &\textbf{\#Test} &\textbf{Avg.length}\\ \midrule
Food  &29,207  &\multirow{2}{*}{34,117} &2,722   &2,747   &\multirow{2}{*}{9.91}     \\ 
Kitchen  &34,886 &\multirow{2}{*}{}  &5,451   &5,659  &\multirow{2}{*}{}     \\
\midrule
Movie  &36,845  &\multirow{2}{*}{58,515} &2,032  &1,978  &\multirow{2}{*}{11.98}    \\ 
Book  &63,937 &\multirow{2}{*}{}  &5,612  &5,730 &\multirow{2}{*}{}   \\ 
\midrule

\end{tabular}
}
\label{sec_exp_tab:dataset}
\end{table}

\textbf{Implementation Details.} For fair comparisons, we adopt the same hyperparameter settings as in previous works~\cite{mifn}. We set the learnable embedding size $q$ to 256, the CLIP-based image and text embedding size $e$ to 512, and the mini-batch size to 256, with a dropout rate of 0.3. The $L_2$ regularization coefficient is selected from $\{0.0001, 0.00005, 0.00001\}$, and the learning rate is chosen from $\{0.001, 0.0005, 0.0001\}$. Training is conducted for 100 epochs using an NVIDIA 4090 GPU, with the Adam optimizer~\cite{adam} to update all parameters.

\begin{table}[t]
\footnotesize
\centering
\caption{Experimental results (\%) on the Food-Kitchen scenario.}
\vspace{0.3cm}
\label{tab:foodkitchen}
\setlength\tabcolsep{4.5pt} 
\resizebox{\columnwidth}{!}{ 
\begin{tabular}{lcccccc}
\toprule
\multirow{2}{*}{Model (Food-Kitchen)} &
\multirow{2}{*}{MRR} &\multicolumn{2}{c}{NDCG} & \multirow{2}{*}{MRR} &\multicolumn{2}{c}{NDCG} \\
\cmidrule(r){3-4}\cmidrule(l){6-7} &(Food)& @5 & @10  &(Kitchen)& @5  & @10   \\
\midrule
GRU4Rec~\cite{gru4rec}   &   5.79  &   5.48   &  6.13 &
 3.06 &  2.55  &  3.10  \\
SASRec~\cite{sasrec}   &   7.30  &   6.90   &  7.79  &
 3.79 &  3.35  &  3.93  \\ 
SR-GNN~\cite{srgnn}   &    7.84  &   7.58  & 8.35   &
  4.01 &   3.47   &  4.13  \\

MIFN~\cite{mifn}  & 8.55  &   8.28  & 9.01   &
  4.09 &   3.57   &  4.29  \\
Tri-CDR~\cite{ma2024triple} &  8.35 &  8.18 &   8.89  &  
4.29  &  3.63 &   4.33  \\
\textbf{Ours HAF-VT} &\textbf{9.16} &\textbf{8.95} &\textbf{9.98}
&\textbf{5.03} &\textbf{4.64} &\textbf{5.35}\\
\bottomrule
\end{tabular}
}
\end{table}

\begin{table}[htbp]
\footnotesize
\centering
\caption{Experimental results (\%) on the Movie-Book scenario.}
\vspace{0.3cm}
\label{tab:moviebook}
\setlength\tabcolsep{4.5pt} % Adjust the column spacing
\resizebox{\columnwidth}{!}{ % Automatically adjust the table to fit the column width
\begin{tabular}{lcccccc}
\toprule
\multirow{2}{*}{Model (Movie-Book)} &
\multirow{2}{*}{MRR} &\multicolumn{2}{c}{NDCG} & \multirow{2}{*}{MRR} &\multicolumn{2}{c}{NDCG} \\
\cmidrule(r){3-4}\cmidrule(l){6-7} &(Movie)  & @5 & @10  &(Book)& @5  & @10   \\
\midrule

GRU4Rec~\cite{gru4rec}   &  3.83 &   3.14 &  3.73  &
 1.68 &  1.34   &  1.52  \\

SASRec~\cite{sasrec}   &   3.79 &   3.23 &  3.69  &
 1.81 &  1.41   &  1.71  \\

SR-GNN~\cite{srgnn}   &  3.85 &  3.27   &  3.78 &
  1.78 &  1.40   &  1.66  \\

PSJNet~\cite{PSJnet}  &  4.63 &  4.06 &   4.76 & 
 2.44  &  2.07 &   2.35  \\

MIFN~\cite{mifn}  &  5.05 &  4.21 &   5.20  &  
2.51  &  2.12 &   2.31  \\
Tri-CDR~\cite{ma2024triple} &  5.15 &  4.62 &   5.05  &  
2.32  &  2.08 &   2.22  \\
\textbf{Ours HAF-VT} &\textbf{6.27}  &\textbf{5.21} &\textbf{5.98}  
&\textbf{2.84} &\textbf{2.45} &\textbf{2.76} \\
\bottomrule
\end{tabular}
}
\end{table}

\subsection{Performance Comparisons}

We evaluate our method against existing state-of-the-art (SOTA) approaches on the ``Food-Kitchen'' and ``Movie-Book'' CDSR scenarios, as summarized in Tab.~\ref{tab:foodkitchen} and Tab.~\ref{tab:moviebook}, respectively. Our approach achieves superior performance, surpassing the current SOTA methods in terms of final evaluation metrics, thereby demonstrating its effectiveness in cross-domain recommendation tasks.

\begin{table}[t]
\centering
\caption{Ablation study on the movie dataset}
\vspace{0.3cm}
\label{tab:ablation}
\begin{adjustbox}{width=1.0\linewidth}
\begin{tabular}{ccccccc}
\toprule
 original-framework & Textual Fusion & Visual Fusion & Hierarchical Attention & MRR \\
\midrule
\checkmark & & &  & 5.03 \\
 \checkmark &  & \checkmark & &  5.63 \\
 
  \checkmark & \checkmark&  & &  5.48 \\
  \checkmark & \checkmark&  & \checkmark&  5.88 \\
  \checkmark &  & \checkmark & \checkmark & 6.08 \\
 \checkmark & \checkmark& \checkmark & \checkmark & 6.27 \\
\bottomrule
\end{tabular}
\end{adjustbox}
\end{table}

\subsection{Ablation Studies}

We conducted an ablation study to systematically evaluate the contributions of our proposed components: \textit{Textual Fusion}, \textit{Visual Fusion}, and \textit{Hierarchical Attention}. As illustrated in Tab.~\ref{tab:ablation}, we first established a baseline using the original framework without any additional modules, achieving an MRR of 5.03. The introduction of \textit{Visual Fusion} alone resulted in a performance gain of 0.6 points, highlighting the importance of incorporating visual signals for cross-domain recommendation tasks. Similarly, adding \textit{Textual Fusion} without hierarchical attention improved the MRR to 5.48, demonstrating the complementary role of textual information in enhancing the model's understanding of user preferences.

When both \textit{Textual Fusion} and \textit{Hierarchical Attention} were integrated, the MRR further increased to 5.88, underscoring the effectiveness of hierarchical structures in capturing domain-specific features. Notably, the combination of \textit{Visual Fusion} and \textit{Hierarchical Attention} achieved an MRR of 6.08, indicating that visual signals and hierarchical modeling synergistically contribute to performance improvements. 

Finally, the full framework, incorporating \textit{Textual Fusion}, \textit{Visual Fusion}, and \textit{Hierarchical Attention}, achieved the highest MRR of 6.27. This result demonstrates the cumulative impact of our proposed modules, with each component playing a distinct yet complementary role in enhancing the model's ability to capture cross-domain user preferences. Overall, our ablation study confirms the effectiveness of the HAF-VT method in advancing CDSR performance.

\section{Conclusion}

In this work, we propose a novel framework, \textit{Hierarchical Attention with Visual and Textual Fusion} (HAF-VT), for Cross-Domain Sequential Recommendation. Departing from conventional methods that focus exclusively on temporal patterns in user interaction sequences, our approach integrates multimodal representations by harnessing the robust visual and textual understanding capabilities of the CLIP model. Specifically, we generate frozen image and text embeddings, which are then fused with a learnable item matrix to enrich item representations. To further enhance the model's ability to capture cross-domain user preferences, we introduce a hierarchical attention mechanism. This mechanism is designed to process sequential information from multiple domains and modalities, enabling the model to discern and leverage distinct features from each source. By combining these innovations, our HAF-VT framework achieves state-of-the-art performance in CDSR, demonstrating the effectiveness of integrating multimodal signals and hierarchical modeling for sequential recommendation tasks.

\section{Acknowledgements}
This work was supported by the National Natural Science Foundation of China (No. 62471405, 62331003, 62301451), Suzhou Basic Research Program (SYG202316), XJTLU REF-22-01-010, and the SIP AI Innovation Platform (YZCXPT2022103).

\bibliographystyle{IEEEbib}

\bibliography{refs}

\end{document}